\documentclass[10pt,twocolumn]{article} 
\usepackage{simpleConference}
\usepackage{times}
\usepackage{graphicx}
\usepackage{amssymb}
\usepackage{url,hyperref}
\usepackage{caption}
\usepackage{subcaption}

\begin{document}

\title{EFFICIENTWORD-NET: AN OPEN SOURCE HOTWORD DETECTION ENGINE BASED ON ONE-SHOT LEARNING}

\author{Chidhambararajan R, Aman Rangapur, Dr. Sibi Chakkravarthy \\
\\
Vellore Institute of Technology-AP, India \\
\\
}


\maketitle
\thispagestyle{empty}

\begin{abstract}
Voice assistants like Siri, Google Assistant, Alexa etc. are used widely across the globe for home automation, these require the use of special phrases also known as hotwords to wake it up and perform an action like "Hey Alexa!", "Ok, Google!", "Hey Siri!" etc. These hotwords are detected with lightweight real-time engines whose purpose is to detect the hotwords uttered by the user. This paper presents the design and implementation of a lightweight, easy-to-implement hotword detection engine based on one-shot learning which detects the hotword uttered by the user in real-time with just one or few training samples of the hotword. This approach is efficient compared to existing implementations because the process of adding a new hotword in the existing systems requires enormous amounts of positive and negative training samples and the model needs to retrain for every hotword. This makes the existing implementations inefficient in terms of computation and cost. The architecture proposed in this paper has achieved an accuracy of 94.51\%.
\end{abstract}

\Section{INTRODUCTION}
With the advent of the Internet of Things (IoT) and home automation, there is a growing need for voice automation in edge devices, but running a heavy Text To Speech (TTS) Engine is too computationally expensive in these edge devices \cite{amodei2015deep}. Instead, we can run engines that need to listen for specific activation phrases called "Hotwords" to perform certain actions since the detection of hotwords is computationally less expensive than full-blown TTS engines \cite{Yang_Jee_Leblanc_Weaver_Armand_2020}.

\begin{figure}
\centering
\includegraphics[width=0.45\textwidth]{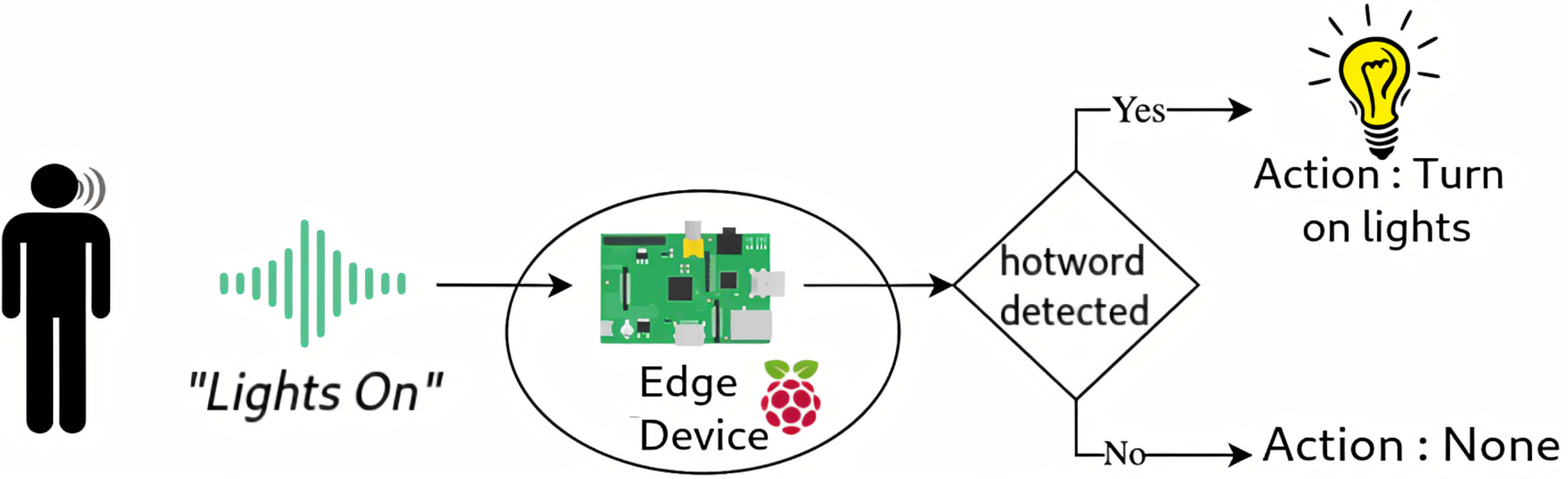}
\caption{Overview of the proposed model.}
\label{FIG:Illustration_of_the_proposed_architecture}
\end{figure}

Lightweight models \cite{Yang_Jee_Leblanc_Weaver_Armand_2020} are trained for detecting these hotwords from audio streams. This is used to save resources from heavy models such as speech recognition from running all day. The Core application for hotword detection is shown in Fig. \ref{FIG:Illustration_of_the_proposed_architecture}.

Convolutional Neural Networks (CNNs) have proven to be the best in analysing image data \cite{Hershey_2017}. Audio files are converted into Log Mel spectrograms where various frequencies are distributed on the Mel scale and plotted as an image \cite{Atsavasirilert_2019,9413528}. This image data is further analysed by the CNNs to get maximum optimacy \cite{Hershey_2017}. The base network is further attached to a Siamese network which learns to output embedding vectors with less distance for similar hotwords and huge distance for dissimilar hotwords. This way a state of the art accuracy is achieved for hotword detection with fewer audio examples of the hotwords. To our knowledge, this is the first attempt to solve the problem of retraining hotwords with one shot / few-shot learning. This approach is highly inspired by Face-Net \cite{Schroff_2015}, one-shot learning deployed for face recognition, which allows us to add a new face to the system without retraining the model \cite{Schroff_2015}. Except for the well-known hotword detection engines with low accuracy, other engines require huge datasets with positive and negative samples for training new hotword and all of these are closed source \cite{Yang_Jee_Leblanc_Weaver_Armand_2020,kalith_2012}. 

EfficientNet \cite{tan2020efficientnet} is one of the most efficient CNN architecture to the date, and the first four blocks of EfficientNet (B0 variant) is chosen for the base model in the Siamese network. To train the Siamese network, positive and negative pairs of audios are given to the network and trained to output 1, 0 for positive pairs and negative pairs respectively to know how similar the pronunciations of words. In the edge device, raw audio is continuously read(with 1 sec time window), converted to Log Mel spectrogram, from which real-time vector embeddings are calculated, these embeddings are compared against a pre-calculated vector embedding of the desired hotword for similarity \cite{Vargas_2020}.

\Section{EXISTING RELATED WORKS}
\label{Sec:Related_Works}
The problem of detecting hotwords from audio streams started ever since the advent of voice-enabled IoT devices \cite{Ooi_2019,Michaely_2017, inproceedings2_Reis_2018,inproceedings3_Todisco_2019,Tom_2018,He_Kaiming_2016,He_2016,Lin_2018,becker2019interpreting,Uitdenbogerd_2004,Tang_2018}. Porcupine \cite{picovoice_alireza} is a closed source hotword detection framework which detects hotwords but requires a commercial licence. It has an accuracy of 94\%. A customizable hotword detection engine called Snowboy can be used to create your own hotwords \cite{Yang_Jee_Leblanc_Weaver_Armand_2020}. It is a closed source project and requires a huge amount of data samples. PocketSphinx \cite{kalith_2012} is a lightweight variation of the CMU-Sphinx \cite{kalith_2012}, an offline Speech-to-Text (STT) engine with low accuracy. 

There are other frameworks such as howl \cite{tang-etal-2020-howl}, these frameworks require large amounts of positive and negative samples of the hotword to train and recognize the new words. Snowboy \cite{chen_yao} and howl’s \cite{tang-etal-2020-howl} accuracy depends on the size of training dataset for each hotword. STT engines can also be used to detect hotwords \cite{amodei2015deep}, existing engines STT \cite{Kubota_patent_2014} have very high accuracy but require a constant internet connection and an expensive subscription to their cloud service to run 24/7. Offline STT open source engines like DeepSpeech (from Mozilla) \cite{amodei2015deep} and Silero \cite{Silero_Models} have good accuracy, yet require lots of on-device resources, hence cannot be run 24/7. Rhino \cite{kenarsari_2018}, an on-device STT engine achieved the best among existing engines, by giving better accuracy and low resource requirement but closed source and requires a commercial license.

Most existing audio processing neural networks employ the usage of Log Mel Spectrograms \cite{Atsavasirilert_2019} and Mel Frequency Cepstral Coefficient \cite{Atsavasirilert_2019} since it conveys a better picture of the audio than conventional audio stream bits, this is due to the representation of changes for various frequencies across the audio streams. The initial development of hotword detection was achieved without noise \cite{Huang_2019} and then deployed with active noise cancellation in real-time \cite{Huang_Yiteng_Wan_Li_2019}. Active noise cancellation is often achieved with hardware first or software first or hybrid approaches. The software first approaches require audio samples recorded from the ambient space. Out of these recorded sample metrics, the maximum amplitude for noise is calculated and used as a threshold for voice activity detection. Acoustic noise reduction is applied over real-time audio with the help of obtained noise only audio samples. This approach is not practical since noise only audio samples recorded from ambient space are required. To circumvent this issue, edge voice assistant’s like Google, Amazon’s Alexa often deploy multi-microphone array systems \cite{Huang_Yiteng_Wan_Li_2019} to gather audio from all directions and separate speech audio with ease. This is a relatively simpler task as devices are surrounded with uniform noise in all directions but speech audio is not uniform in all directions. This allows speech audio to be separated from noise audio. The separated speech audio is of high quality thereby helping in achieving low False Acceptance Rates (FAR).

In a hybrid approach, hardware-based noise cancellation is utilized and the audio processing neural networks are often trained with noise, this allows the system to achieve exceptionally low FAR \cite{Huang_Yiteng_Wan_Li_2019}.

Existing audio processing neural networks are designed as audio classification neural networks \cite{Hershey_2017}. The network needs a lot of initial layers to understand the audio fragment. Later, half a dozen of layers are required for the network to understand the logic of classification. These extra layers dedicated for classification requires additional computational time thereby crippling the model while running it on real-time edge devices. In this paper, this problem was resolved using EfficientNetB0 as a base network with one-shot learning. In one-shot learning, the network requires a lot of layers to understand audio samples, but the need for additional layers to understand classification is waived off, therefore, resulting in faster inference in the edge devices.

As mentioned, the existing lightweight models \cite{Yang_Jee_Leblanc_Weaver_Armand_2020,kalith_2012} are binary classifiers that need to be retrained with a huge number of negative and positive samples for a new hotword, this results in expensive and inefficient gathering of datasets \cite{kalith_2012}. These networks needs to be retrained for newer hotwords again. Also, these engines are closed source where there is no scope of development in the future and users need to spend lots of money for the useage. Hardware first and a hybrid approach are applicable in the scenarios where the edge device’s hardware specifications are under the control of developers like edge voice assistants like Alexa \cite{Huang_2019}. But this approach doesn’t work well when the edge device is not designed by developers \cite{Huang_Yiteng_Wan_Li_2019}. Hence, there is a need for audio processing neural networks to be trained with very high amounts of noise to work robustly without need of hardware interventions.

EfficientWord-Net is an open source engine that solves the process of retraining the model for new hotwords by eliminating the requirement of huge datasets. It works efficiently with the audio samples with decent noise added in the background with a great inference time on small devices like Raspberry Pi. Moreover, our system outperformed previous existing approaches in terms of accuracy and inference.

\Section{ONE-SHOT LEARNING/SIMILARITY LEARNING}
\label{Sec:oneshotlearning}
One of the demanding situations of face recognition/hotword recognition is to achieve performance with fewer samples of the target, which means, for maximum face recognition programs, model should recognize a person given with the aid of using one photo of the man or woman's face. Traditionally deep learning algorithms do not work well with the simplest one training example or one data point for a class. In one-shot learning, a model learns from one sample to apprehend or recognize the person, and the industry needs most face recognition models to use this due to the fact a company has one or few images of every of their personnel in the database. 


Similarity learning is a type of supervised learning where a network is trained to identify the similarity between 2 data points of the same class instead of classification or regression. The network is also trained to learn the dissimilarity between 2 data points of different sources. This similarity is used to determine whether an unlabelled data point belongs to the same class or not.

When 2 images are fed to a neural network to learn the similarity between them (inputs two images and outputs the degree of difference between the two images), the output would be a small number if the two images are of the same person. And if they are of two different people, the output would be a large number. However, the use of different types of functions keeps the value between 0 and 1. A hyper parameter($\tau$) is used as a threshold if the degree of difference is less than the threshold value, these pictures belong to the same person and vice versa. Similarly, in this paper, a threshold value of 0.2 is defined, if the degree of difference between the two hotwords is less than 0.2, those two hotwords are the same, else they are different.
\\

d\emph{(hotword1, hotword2)} $=$ degree of difference between hotwords.

d\emph{(hotword1, hotword2)}  $ < \tau$, both hotwords are same.

d\emph{(hotword1, hotword2)}  $ > \tau$, hotwords are different. 
\\

This approach of feeding 2 different images to the same convolutional network and comparing the encodings of them is called a Siamese Network \cite{Vargas_2020}.

\begin{figure} 
\centering
	\includegraphics[width=0.5\textwidth]{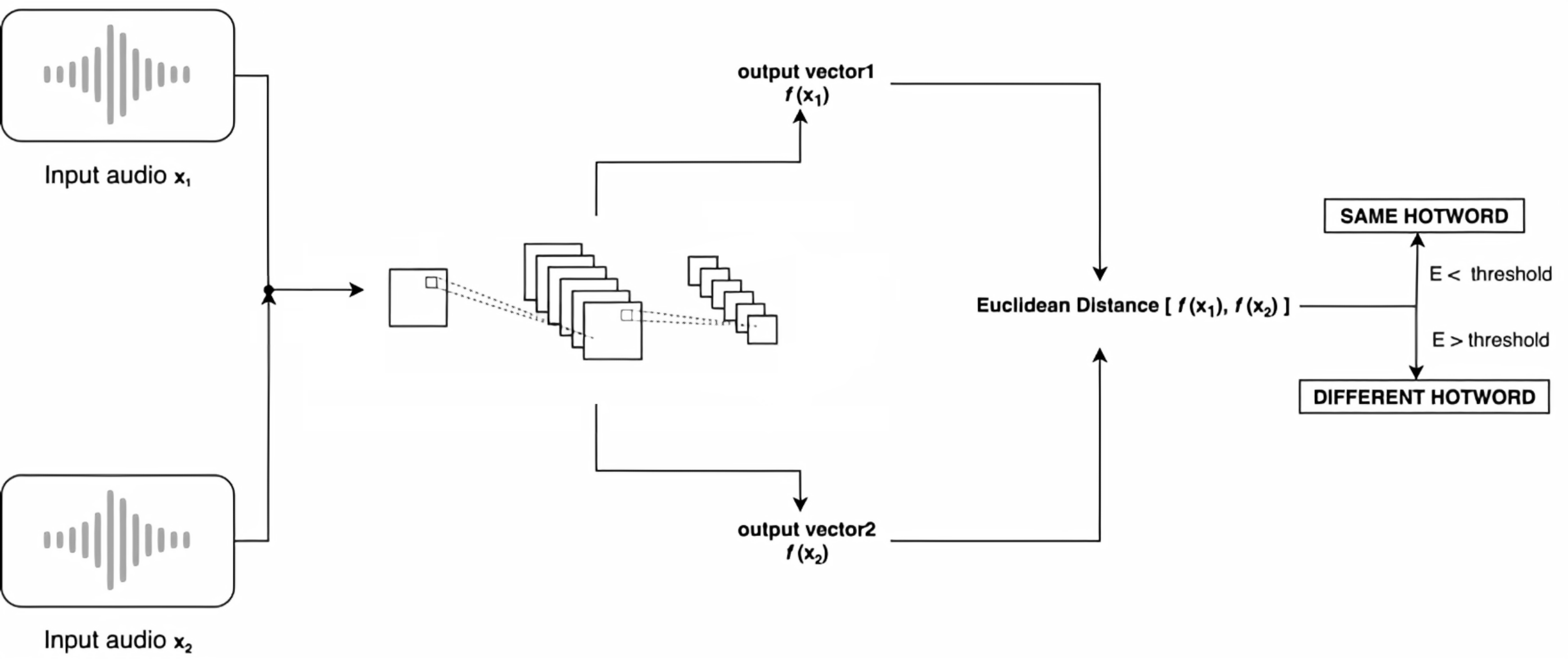}
\caption{One-shot learning architecture for hotword detection}
\label{FIG:OneshotLearningArchitecture}
\end{figure}

In Figure. \ref{FIG:OneshotLearningArchitecture}, two raw audio segments are fed to the same neural network, output encodings of input audios are captured, Euclidean distance between these two vectors are calculated, if the Euclidean distance is less than the threshold value, the hotwords in the audio are same and vice versa.

\Section{PROPOSED METHOD}
\label{Sec:efficientword-net}
\subsection{Preparation Of Dataset}
The dataset used for training the network is homegrown artificially synthesized data that was made with naturally sounding neural voices from Azure Cloud Platform and Siri. Furthermore, the voices were selected to include all available voice accents with the respective countries to ensure better performance across accents and gender. Hotword detection does not rely on a word’s meaning but only on its pronunciation. So, to generate the audio, a pool of words is selected in which each word sounds unique compared to the other words in the pool. Google’s 10000-word list \cite{kaufman_bathman_myers_hingston} is used while training the model, and these words were converted to respective phoneme sequences. The sequences were checked for similarity, the words which shared $\ge$80\% of the same phoneme sequence were removed to ensure low similarity in pronunciation among the word pool. The word pool was converted to the audio pool with the above mentioned text-to-speech services. The generated audio is combined with noises such as traffic sounds, market place sounds, office and room ambient sounds to simulate a naturally collected dataset. For each word, 5 such audio samples were generated in which each sample had a randomly chosen voice and randomly chosen background noise to ensure variety. The randomly chosen noises are then combined with the original audio with a noise factor randomly chosen between 0.05 to 0.2 (noise factor is a fraction of noise’s volume in the resulting audio). A sampling rate of 16000 Hz was chosen since audio quality below 16000 Hz became very poor. Finally, the audios are converted to Log Mel Spectrograms.

Two audios generated from the same word are chosen to make the true pair and two audios generated from two different words are chosen to make the false pair. The total number of true and false pairs generated in this method were 2694. 80\% of the data was split for training, and remaining 20\% was used as testing data to eliminate over-fitting. The network is fed with the true pair and false pairs to output a higher similarity score for true pairs and lower similarity score for false pairs.

\subsection{Network Architecture}
\begin{figure}
\centering
\includegraphics[width=0.5\textwidth]{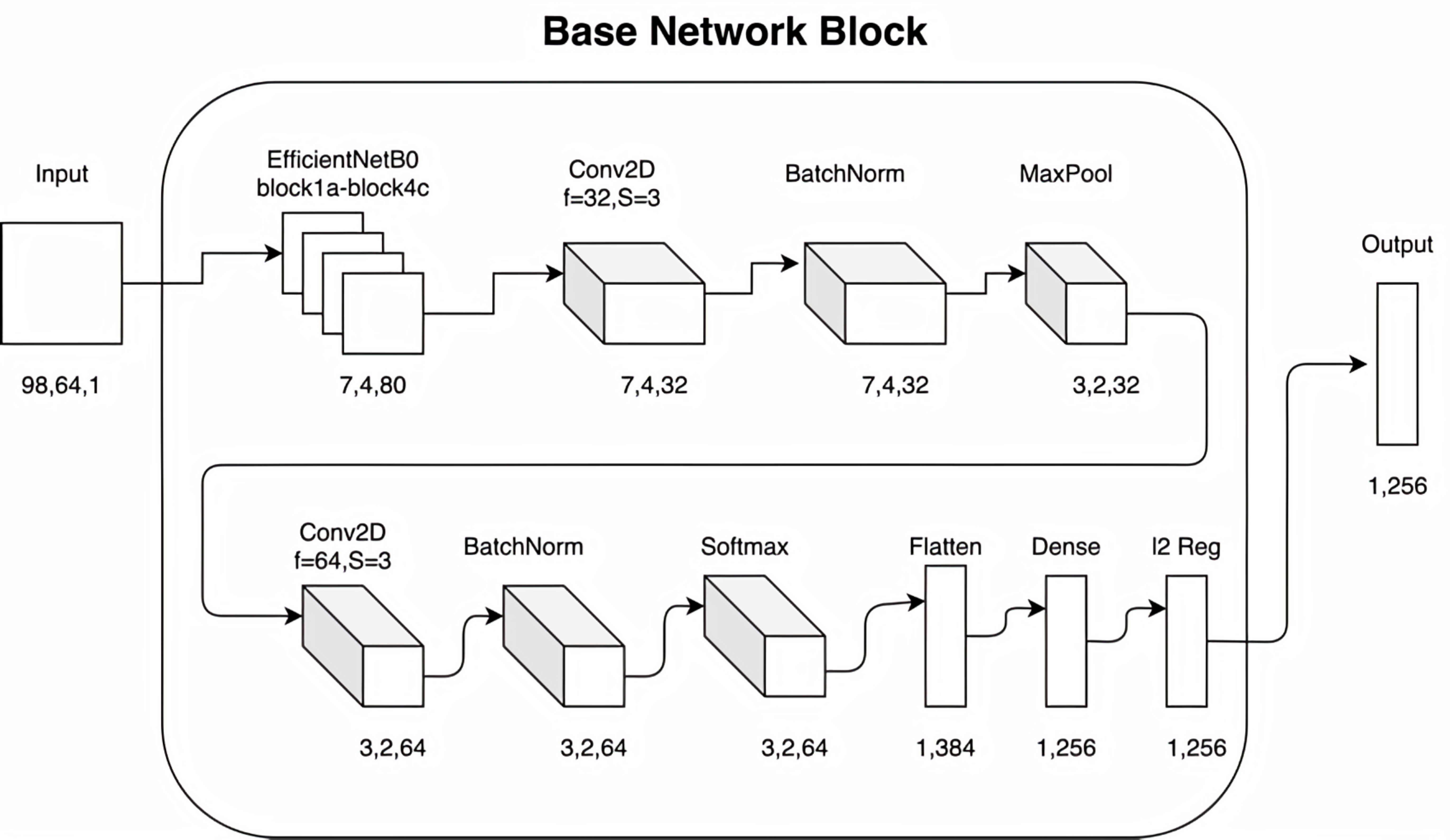}
\caption{Base network Block.}
\label{FIG:Base_network_Block}
\end{figure}

The input for each base network is 98x64x1 (Refer Fig. \ref {FIG:Base_network_Block}. The base network of the model is made up of the first four blocks of EfficientNetB0 architecture \cite{tan2020efficientnet}. The output from the EfficientNet layers is processed further with Conv2D layer with 32 filters and 3 stride values, processed by batch normalization and max pool layer, this is fed to a similar stack of layers. This is done to reduce the number of feature points efficiently. Finally, the output from the convolutions is flattened and attached to the dense layer with 256 units followed by L2 regularization to give the reduced vector approximation of the input.

A true pair or a false pair in the dataset is fed to two parallel blocks of the base network, where these parallel blocks share the weights. Euclidean distance between corresponding output vector pairs is calculated. Table. \ref{TBL:Euclidean_Distance_and_Similarity_score} shows the Euclidean distance mapped to similarity score to the scale 0-1.0. 

\begin{table}
\centering
\caption{Euclidean Distance and Similarity score}
\label{TBL:Euclidean_Distance_and_Similarity_score}
\begin{tabular}{ p{2cm} p{2cm}   p{2cm}}
 \hline \hline
Euclidean Distance  &Similarity Score \\
 \hline
0   &1.0 \\
 $<\tau$  &1.0 - 0.5 \\
$\tau$ &0.5 \\
 $>\tau$	&0.5 - 0 \\
\hline
\end{tabular}
\end{table}%

\begin{figure}
\centering
\includegraphics[width=0.5\textwidth]{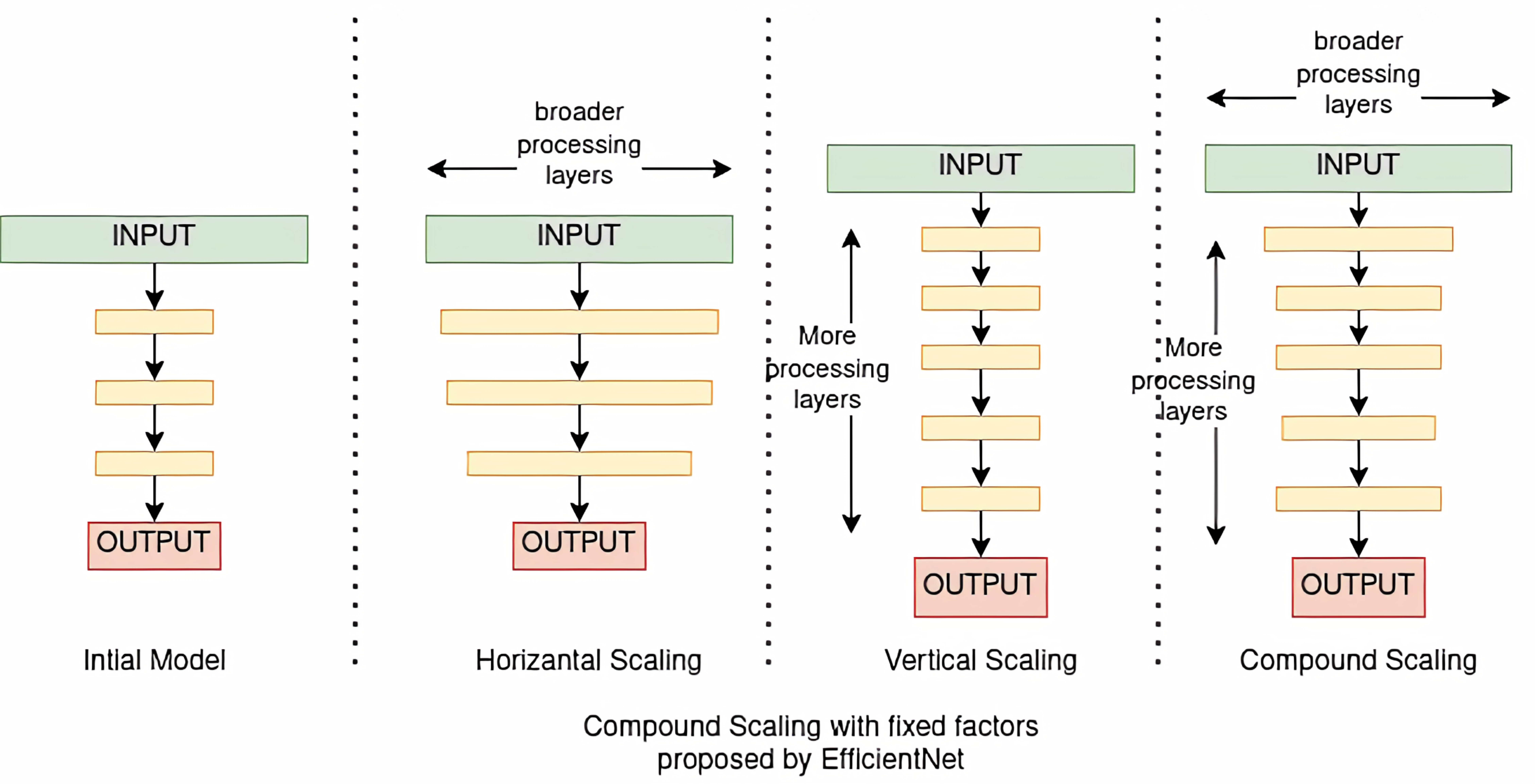}
\caption{Model Scaling in EfficientNet.}
\label{FIG:Model_Scaling_in_EfficientNet}
\end{figure}

\subsection{Training Parameters and Loss Function}
The problem of analysing raw audio to examine images is resolved by converting the audio into Log Mel spectrograms images, this helps the neural network by allowing it to directly analyse frequency distribution over time, learning first to identify different frequencies and analyse them.

These generated images were fed to a convolution neural network that follows the EfficientNet Architecture (Refer Fig. \ref{FIG:Model_Scaling_in_EfficientNet}). It is a convolutional neural network architecture and a scaling method that uniformly scales all dimensions of depth/width/resolution using a compound coefficient. 

This network architecture was chosen since the accuracy was similar to that of ResNet which held the previous state of the art top5 accuracy of 94.51\%. Moreover, the number of parameters in ResNet was 4.9x higher than EfficientNetB0’s parameter count, thereby making it computationally more efficient than ResNet. Only 4 blocks of EfficientNetB0 were taken for the base network, the output was further attached to a Conv2D block, which was later flattened and l2 normalized to give the output vector.

Conventional Siamese neural networks use triplet loss where a baseline (anchor) input is compared to a positive(true) input and a negative(false) input with vector distance calculation metrics such as Euclidean distance, cosine distance, etc.
\\

Triplet Loss Function =
\(\max\left( {\| f\left( x^{a} \right) - f\left( x^{p} \right) \|}^{2} - {\| f\left( x^{a} \right) - f\left( x^{n} \right) \|}^{2},0 \right)\)
\\

\(x^{a}\) is an~\emph{anchor}~example.

\(x^{p}\) is a~\emph{positive}~example that has the same identity as the anchor.

\(x^{n}\) is a~\emph{negative}~example that represents a different
entity.
\\

The triplet loss function makes the neural network minimize the distance from the baseline(anchor) input to the positive (true pair), and maximize the distance from the baseline(anchor) input to the negative (false pair). It is also equipped with a threshold, which forces the network to reduce the distance between true pairs below the threshold and false pairs beyond the threshold.

The calculated distance between the pair is sent through a function F(x) which outputs close to 1 when distance is low and 0 when distance is high, thereby making the function give a score close to 1 for similar pairs and close to 0 for dissimilar pairs. This was done so that the network will be able to tell similarities between a pair of samples in terms of percentage. 
\[F\left( x \right) = 1 - \frac{x^{4}}{({\tau}^{4}+x^{4})}\]

Here x is the calculated distance between the vectors, this function gives 1 when x is 0, gradually reduces to 0.5 when x = $\tau$ (threshold,t[in the above equation]) and eventually to zero. The function is symmetric making f(x) go to zero when x\textless0, with Euclidean distances $\ge$ 0.

For each true pair, the ground truth was set 1 and for each false pair the ground truth was set to 0, this allowed us to treat the problem as a binary classification problem and easily apply binary cross-entropy loss function while training the engine.
\\

\textbf{Binary cross-entropy loss is defined by}

\[loss = - \frac{1}{N}\sum_{i = 1}^{N}y_{i} \bullet log\left( p\left( y_{i} \right) \right) + \left( 1 - y_{i} \right) \bullet log\left( 1 - p\left( y_{i} \right) \right)\]

\subsection{Optimization Strategies}

\subsubsection{Log Mel Spectrogram:~}

Many audio processing neural networks directly process the audio with several Conv1D layers stacked on top of each other to make the network understand the audio and it has a drawback. The network would first have to understand the concept of frequency and check for the distribution of various frequencies across the audio. This forces the network to allocate its initial Conv1D layers to understand the concept of frequency and learn to check for the distribution of various frequencies across the audio, which would be further analysed by the next layers to make sense of the audio. These additional preprocessing layers can be skipped by going for a Log Mel Spectrogram (a heatmap distributing various frequencies across the audio).

Since the network is being directly fed with the distribution of various frequencies across the audio, the network can allocate all of its resources to make sense of the audio directly, thereby enhancing the accuracy of the system. This method is largely inspired by Google's TensorFlow magenta (a set of audio processing tools for TensorFlow).

\begin{figure}
\centering
\includegraphics[width=0.5\textwidth]{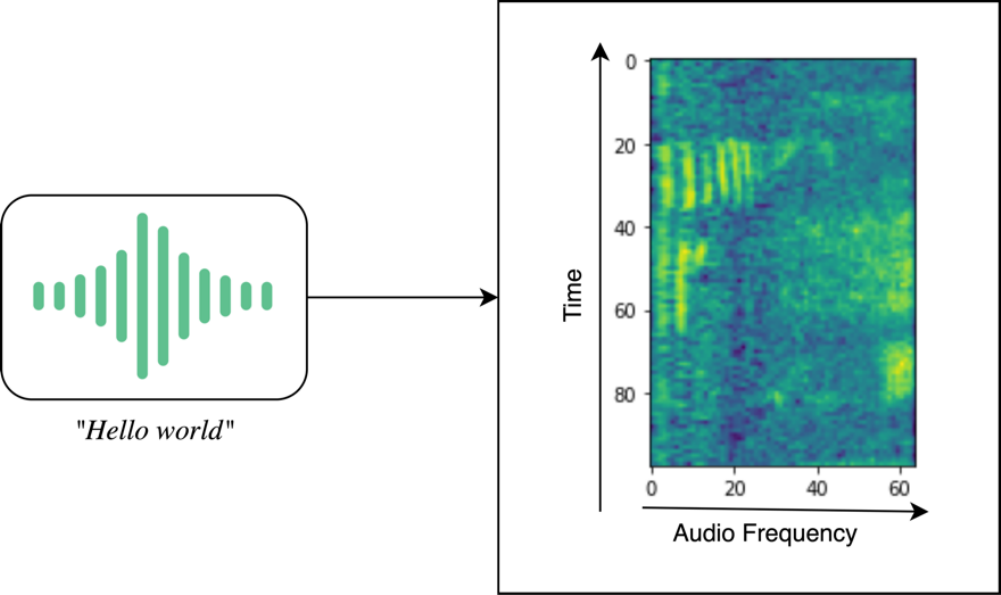}
\caption{Sample audio converted to Log Mel Spectrogram.}
\label{FIG:SampleaudioconvertedtoLogMelSpectrogram}
\end{figure}

\subsubsection{L2 Regularization to an output vector of base network:}

Initially, the network was trained with no L2 regularization, the accuracy was never able to cross 89.9\%, we have tried various techniques out of which adding L2 regularization to the output vector of the base network gave the best performance. With that added, our network was able to reach 95\% accuracy with noise. This performance increase was due to the fact that L2 regularization confined the output vector to the surface of a 256-dimensional unit sphere at the origin; without this confinement, it was very easy for the network to give huge distances for false pairs. Hence, the network eventually became biased, worked well only for false pairs and didn't give small distances for true pairs resulting in the performance drop.

\begin{figure*} [htbp]
\begin{subfigure}{0.5\textwidth}
  \centering
  \includegraphics[width=.7\linewidth]{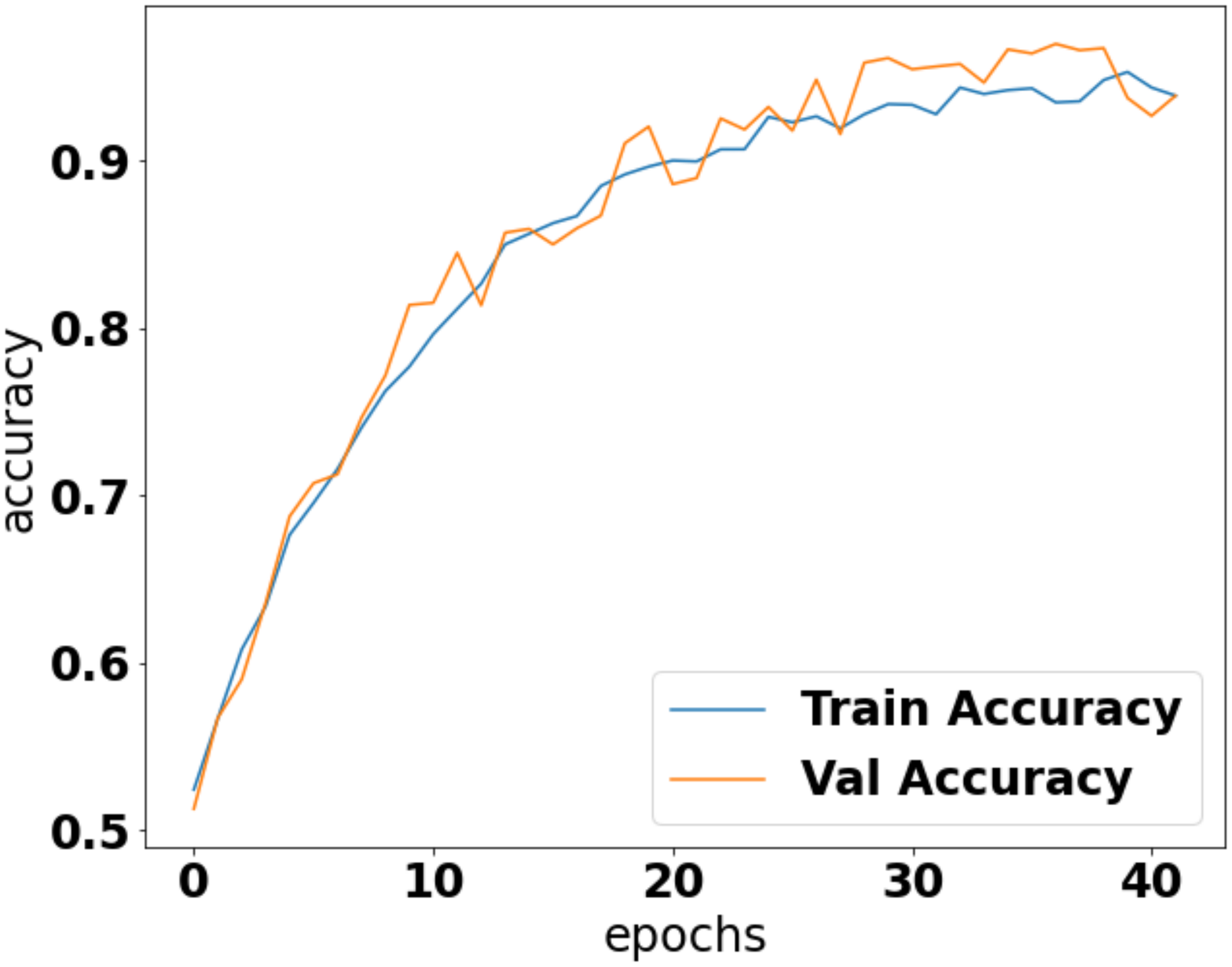}  
  \caption{Model training(Epochs vs Accuracy) with noise}
  \label{FIG:EpochvsAcc1}
\end{subfigure}
\begin{subfigure}{.5\textwidth}
  \centering
  \includegraphics[width=.7\linewidth]{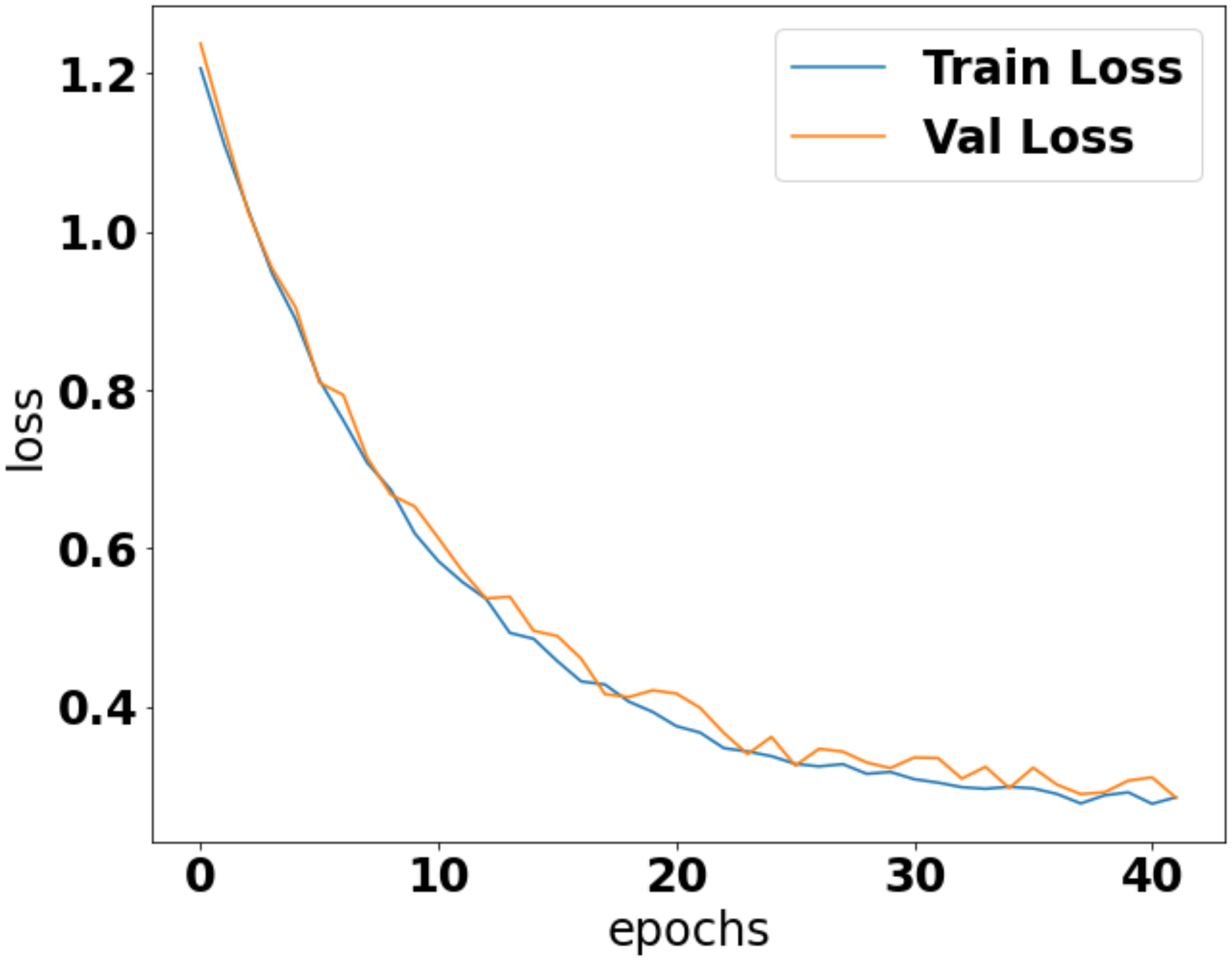}  
  \caption{Model training(Epochs vs Loss) with noise}
  \label{FIG:EpochvsLoss1}
\end{subfigure}\\
\begin{subfigure}{.5\textwidth}
  \centering
  \includegraphics[width=.7\linewidth]{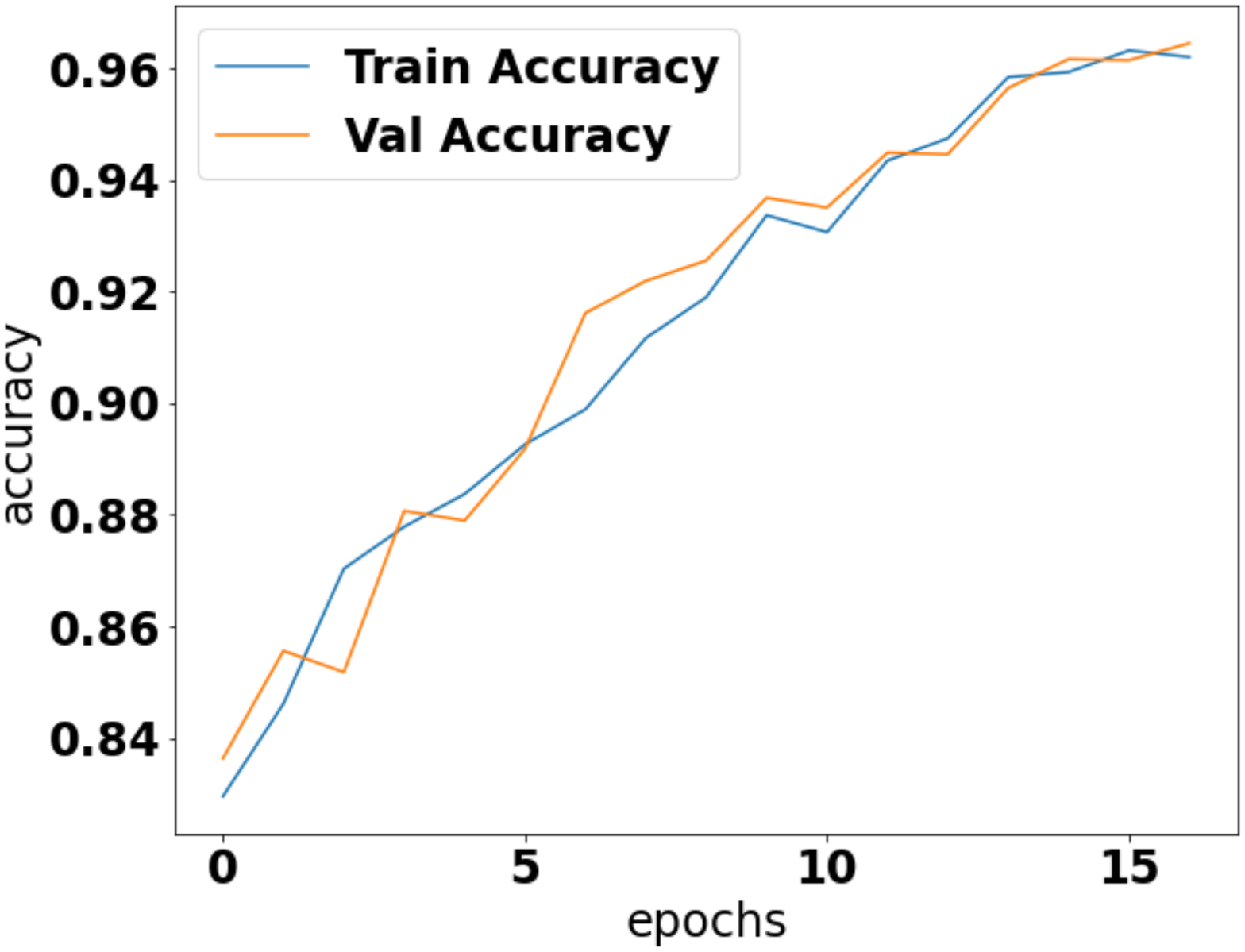}  
\caption{Model training(Epochs vs Accuracy) without noise}
 \label{FIG:EpochvsAcc2}
  \end{subfigure}
\begin{subfigure}{.5\textwidth}
  \centering
  \includegraphics[width=.7\linewidth]{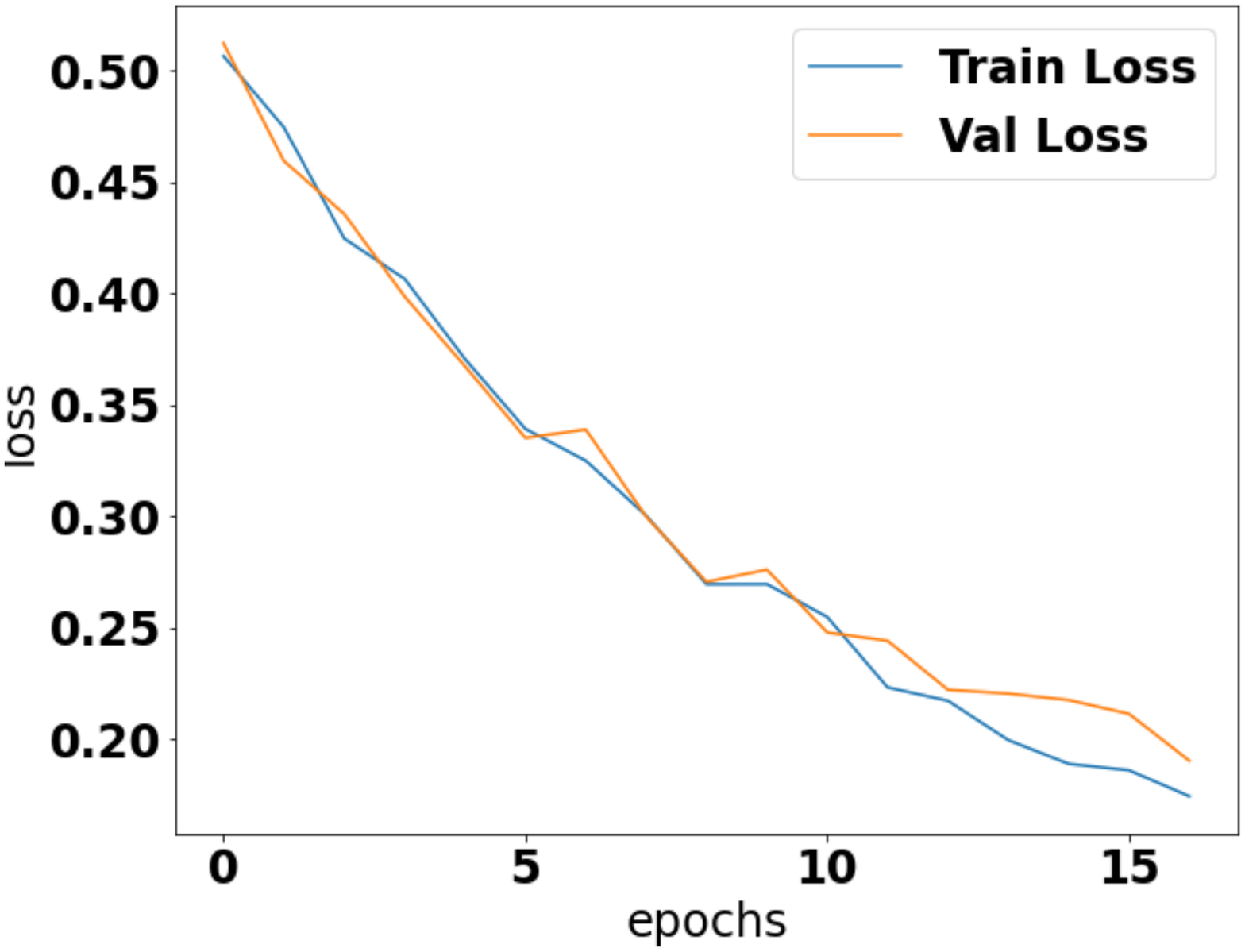}  
\caption{Model training(Epochs vs Loss) without noise}
  \label{FIG:EpochvsLoss2}
  \end{subfigure}
\caption{Model Accuracy and Loss}
\label{fig:model_accuracy_loss}
\end{figure*}

\Section{RESULTS}
\label{Sec:results}
The network was trained with noise using a batch size of 64, 75 training steps per epoch for 42 epochs beyond which the training loss started to oscillate. Adam optimization algorithm was used with an initial learning rate of 1e-3, this learning rate was reduced by a learning rate scheduler which checks for lack of decrease in training loss for 3 epochs and reduces the learning rate by a factor of 0.1. The reduction of the learning rate was stopped when the learning rate reached an absolute minimum of 1e-5. Early stopping was scheduled with patience of 6 which stops the training process when the training loss starts to oscillate in more than 6 epochs resulting in the maximum validation accuracy of 94.51\%. Fig. \ref{FIG:EpochvsAcc1} shows the training graph accuracy/loss vs epochs with noise.

The maximum validation accuracy reached without noise is 96.8\%, this accuracy was achieved by retraining the previous model with the same hyperparameters and noise factor = 0 for 14 epochs. Fig. \ref{FIG:EpochvsAcc1} - Fig. \ref{FIG:EpochvsLoss2} shows the training graph for epoch/accuracy vs loss with and without noise.  Since the real-time audio is chunked into 1-sec windows with 0.25-sec hop length for inference, the base network was found to perform inference in 0.08 seconds in Raspberry Pi 4. Hence, the model will be able to perform inference from real-time audio streams in edge devices with no latency issues. Fig. \ref{FIG:EpochvsLoss1} shows the noise training graph for loss vs epochs.

\begin{figure*}[htbp]
\begin{subfigure}{.5\textwidth}
  \centering
  \includegraphics[width=.6\linewidth]{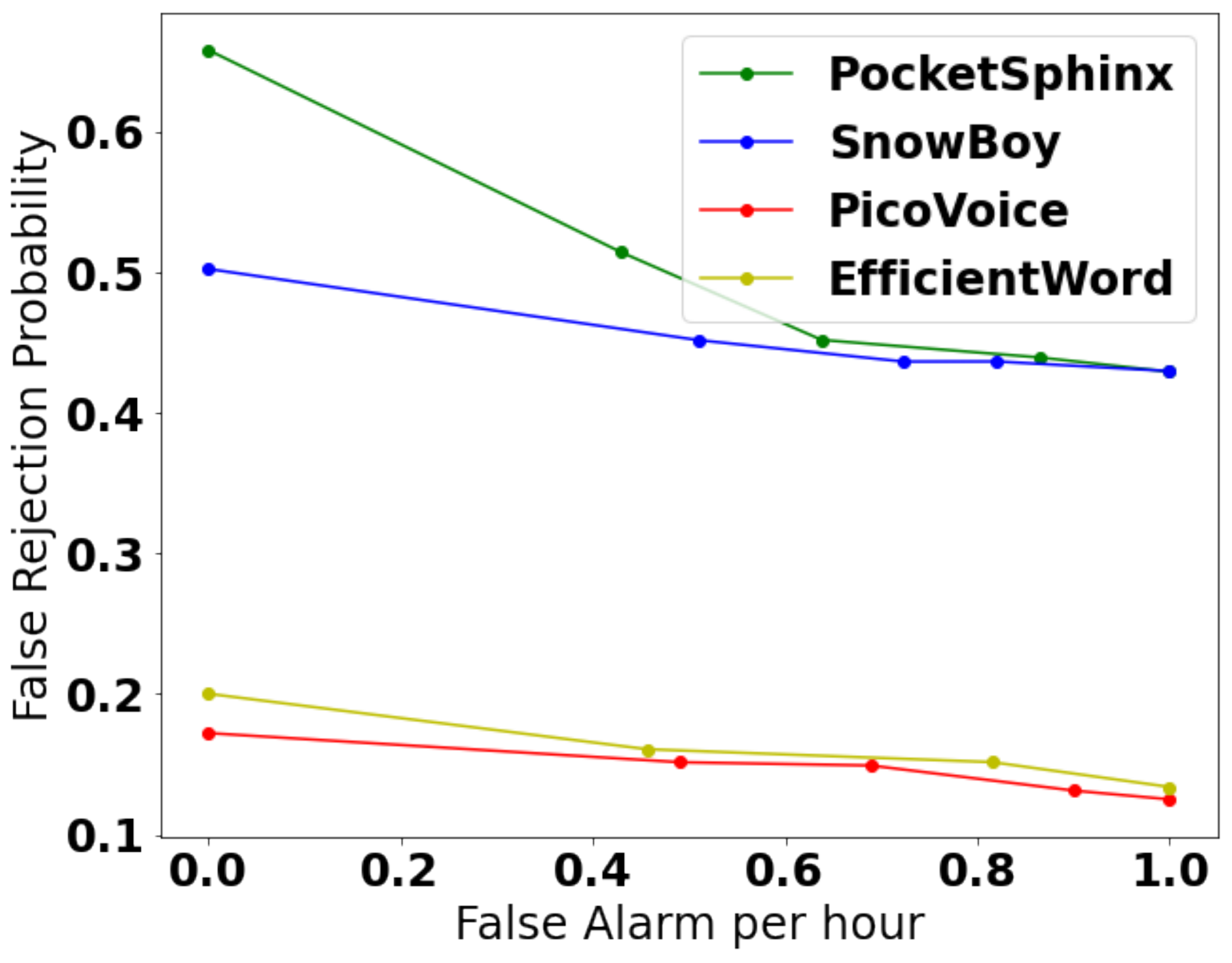}  
  \caption{FRP vs FAR for hotword Alexa}
  \label{FIG:FRPvsFARalexa}
\end{subfigure}
\begin{subfigure}{.5\textwidth}
  \centering
  \includegraphics[width=.6\linewidth]{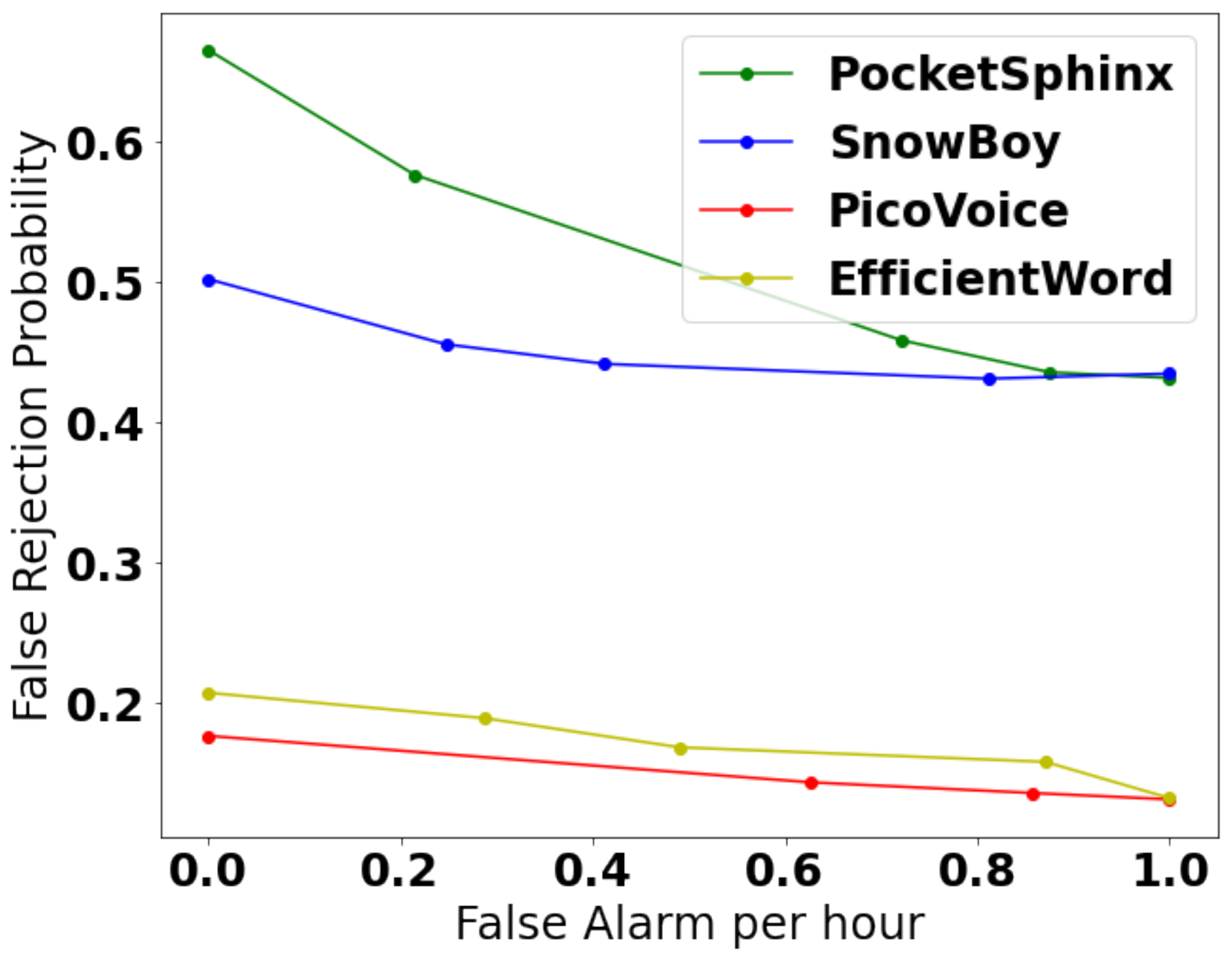}  
  \caption{FRP vs FAR for hotword computer}
  \label{FIG:FRPvsFARcomputer}
\end{subfigure}\\
\begin{subfigure}{.5\textwidth}
  \centering
  \includegraphics[width=.6\linewidth]{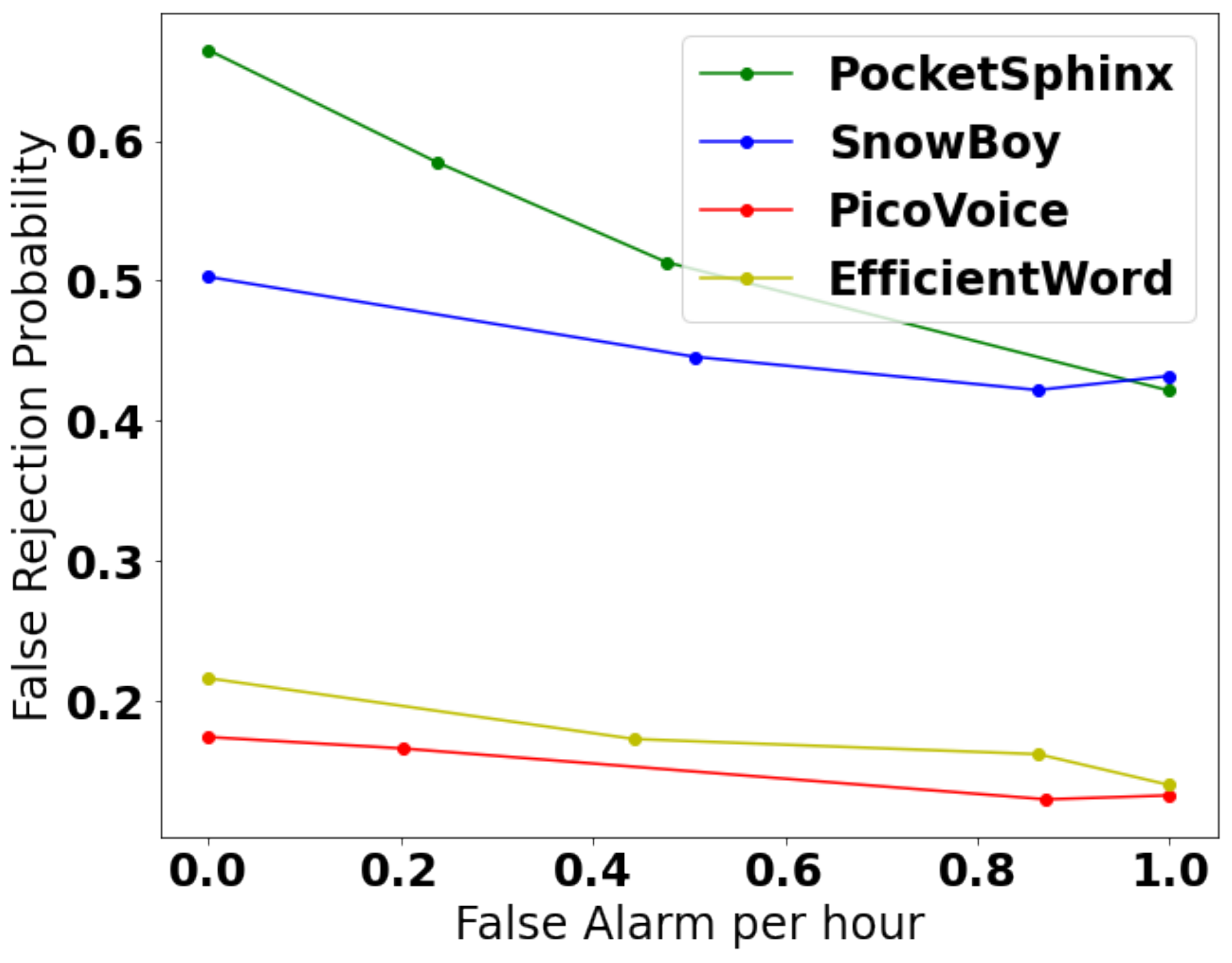}  
\caption{FRP vs FAR for hotword people}
 \label{FIG:FRPvsFARpeople}
  \end{subfigure}
\begin{subfigure}{.5\textwidth}
  \centering
 \includegraphics[width=.6\linewidth]{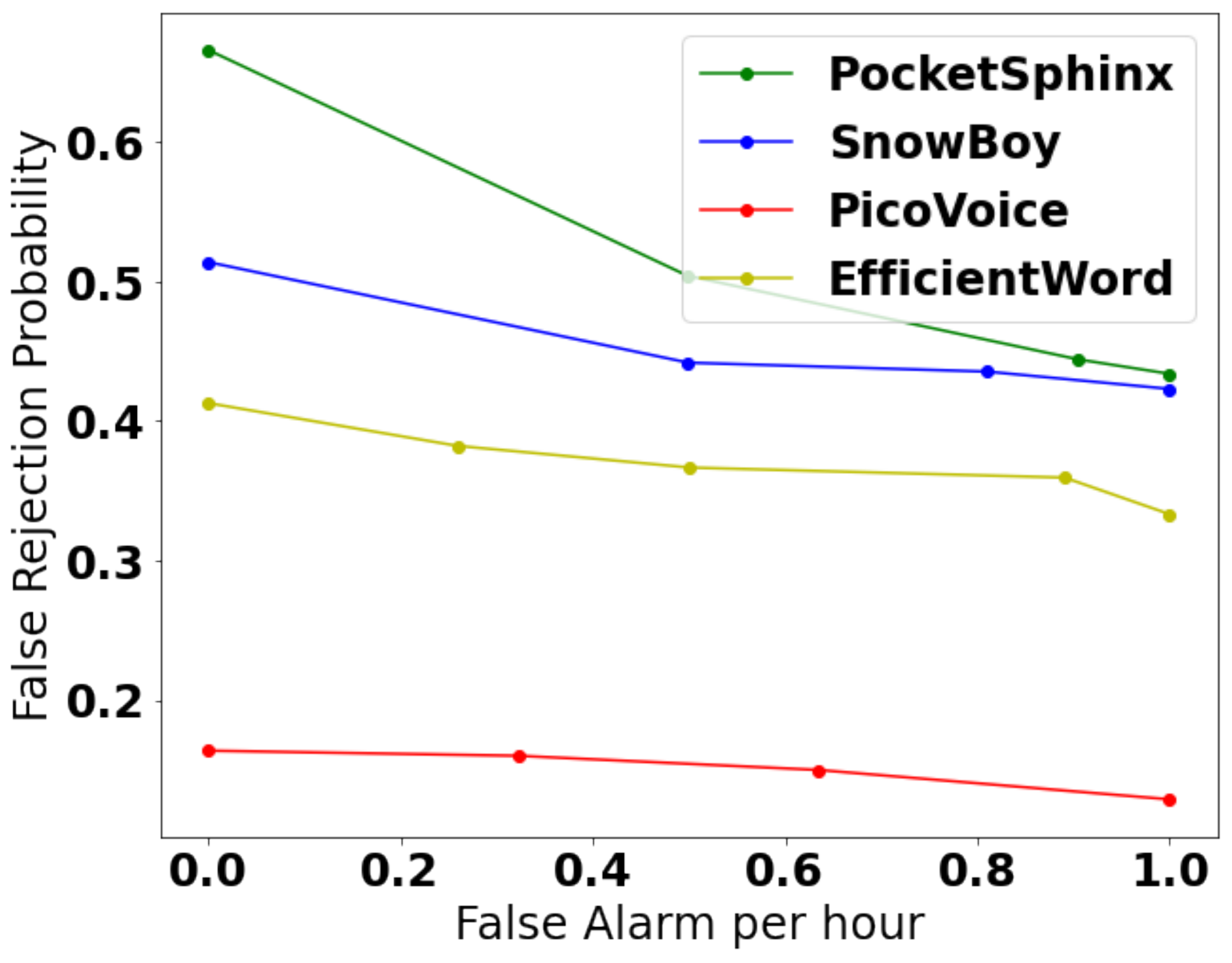}  
\caption{FRP vs FAR for hotword restaurant}
  \label{FIG:FRPvsFARrestaurant}
  \end{subfigure}
\caption{FRP vs FAR for different Hotwords}
\label{fig:frp_vs_far}
\end{figure*}

\begin{table}
\centering
\caption{Model benchmarks}
\label{TBL:Model_benchmarks}
\begin{tabular}{ p{2cm} p{2cm}   p{2cm}}
 \hline \hline
Model   &Accuracy &Inference \\
 \hline
Efficientword-Net (Current paper) &94.51\%   &0.071ms \\
Porcupine \cite{picovoice_alireza} &94.78\% &0.02ms \\
PocketSphinx \cite{kalith_2012} &54.23\% &0.076ms \\
Snowboy \cite{Yang_Jee_Leblanc_Weaver_Armand_2020} &88.43\% &0.091ms \\
\hline
\end{tabular}
\end{table}%

After performing significance test, the resulting trained model was benchmarked with other hotword detection systems on Raspberry Pi 3 clocked at 1.2GHz (4 core) and displayed in the Table. \ref{TBL:Model_benchmarks} and was found to outperform existing closed source models in terms of accuracy by a small level.

\begin{figure*}[htbp]
\begin{subfigure}{.5\textwidth}
  \centering
  \includegraphics[width=.5\linewidth]{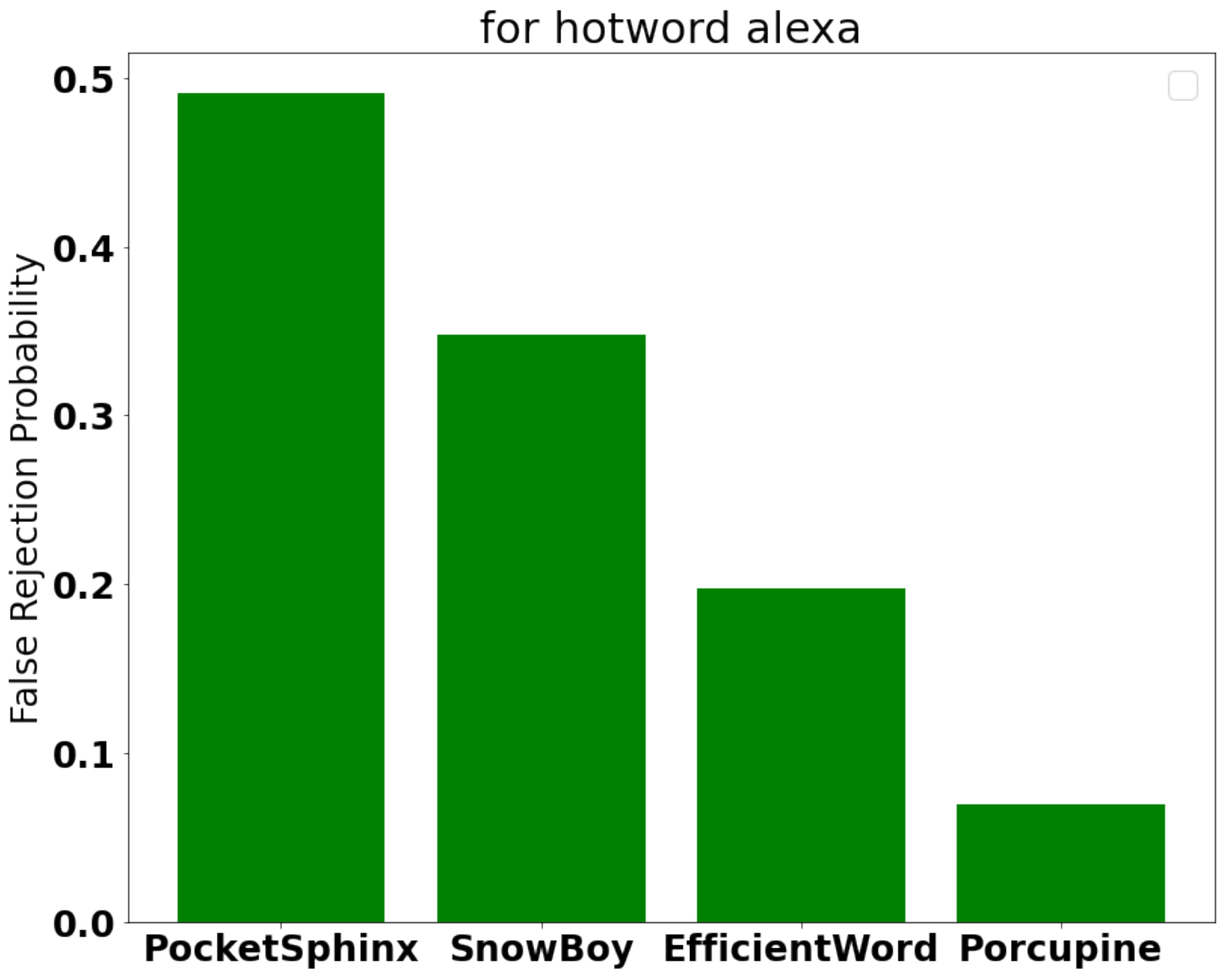}  
  \caption{Bar graph for FRP for hotword Alexa}
  \label{FIG:BarGraphAlexa}
\end{subfigure}
\begin{subfigure}{.5\textwidth}
  \centering
  \includegraphics[width=.5\linewidth]{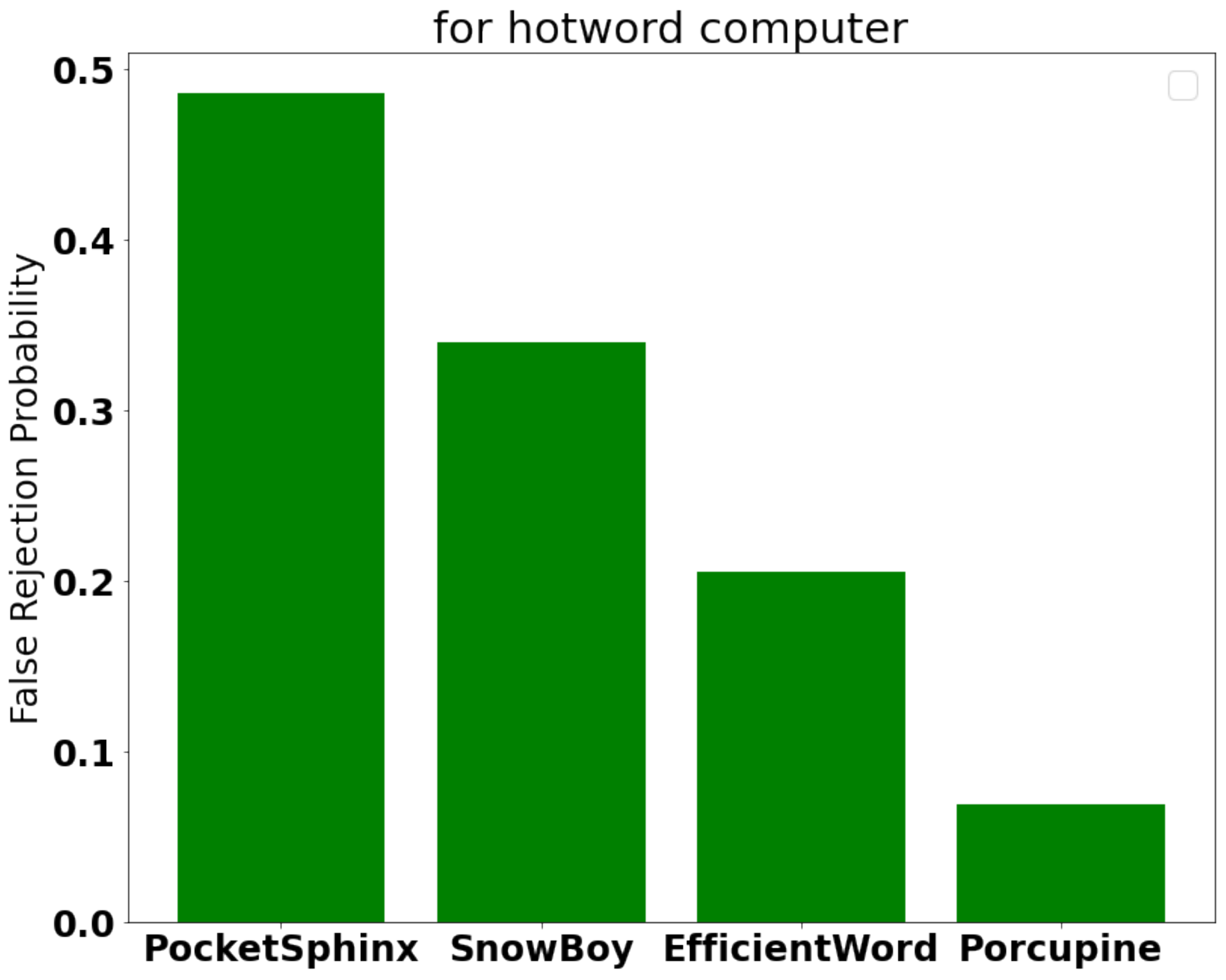}
  \caption{Bar graph for FRP for hotword computer}
  \label{FIG:BarGraphComputer}
\end{subfigure}
\\
\begin{subfigure}{.5\textwidth}
  \centering
  \includegraphics[width=.5\linewidth]{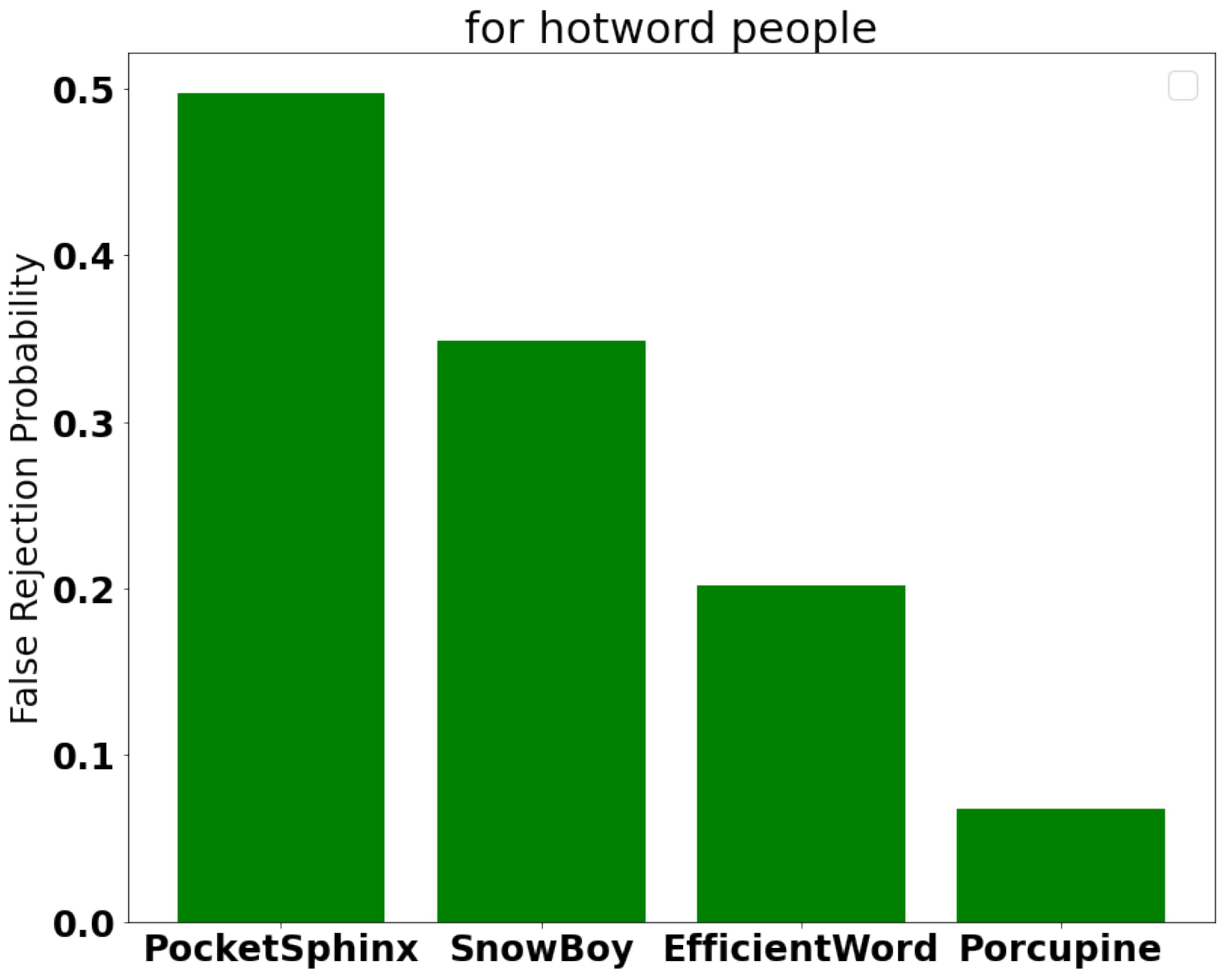}  
\caption{Bar graph for FRP for hotword people}
\label{FIG:BarGraphPeople}
  \end{subfigure}
\begin{subfigure}{.5\textwidth}
  \centering
 \includegraphics[width=.5\linewidth]{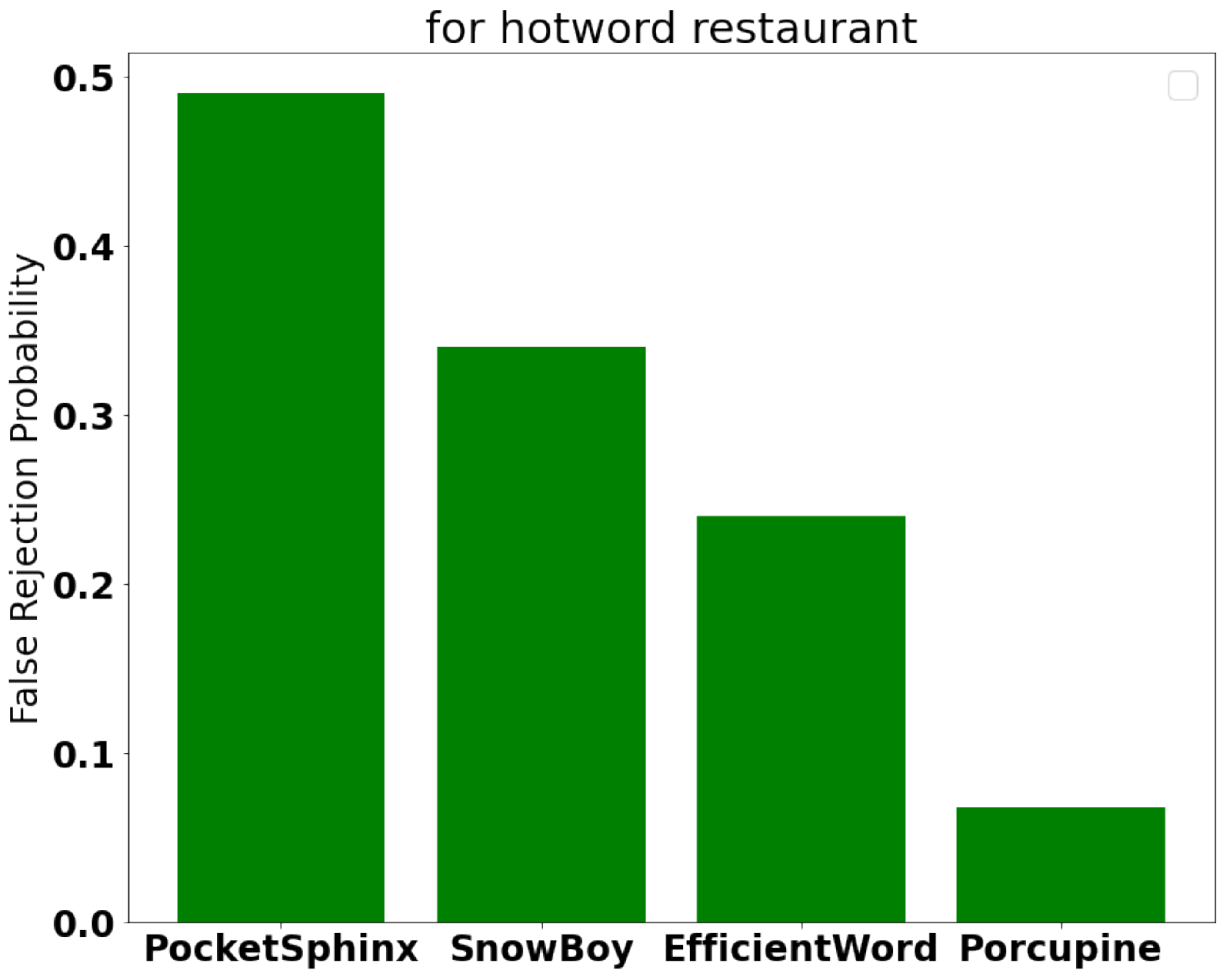}  
\caption{Bar graph for FRP for hotword restaurant}
 \label{FIG:BarGraphRestaurant}
  \end{subfigure}
\caption{Bar graphs for FRP of different Hotwords}
\label{fig:bar_graphs}
\end{figure*}

For a given sensitivity value, False Rejection Rate (FRP) – (True Negatives) is measured by playing a set of sample audio files which include the utterance of the hotword, and then calculate the ratio of rejections to the total number of samples. False Acceptance Rate (FAR-False Positives) is  measured by playing a background audio file which must not include any utterance of the hotword calculated by dividing the number of false acceptances by the length of the background audio in hours. Figures Fig. \ref{FIG:FRPvsFARalexa} - \ref{FIG:FRPvsFARrestaurant} and \ref{FIG:BarGraphAlexa} - \ref{FIG:BarGraphRestaurant} illustrate FRP vs FAR and model performance against existing implementations for various hotwords. All these hotwords are not included in the training dataset.

\Section{CONCLUSION}
\label{Sec:conclusion}

In this paper, we proposed a one-shot learning-based hotword detection engine to solve the problem of retraining and huge dataset requirements for each new hotword with good inference time on light-weight devices. To achieve the same we implemented Siamese neural network architecture with an image processing base network made with EfficientNet, which processes the Log Mel spectrograms of the respective input audio samples. Moreover, this network could also be repurposed for phrase detection's where a program needs to check for the occurrence of a specific sentence removing the requirement of heavy speech-to-text engines in edge devices. Such an engine can allow the end-users to set custom hotwords in their systems with minimal effort.

\Section{ACKNOWLEDGMENTS} 

This research is carried out at Artificial Intelligence and Robotics (AIR) Research Centre, VIT-AP University. We also thank the management for motivating and supporting AIR Research Centre, VIT-AP University in building this project.

\bibliographystyle{abbrv}
\bibliography{output}
\end{document}